\documentclass[]{gmaart2022}

\renewcommand{\varProcName}{Proc. 35. Workshop Computational Intelligence}
\renewcommand{\varLocation}{Berlin}
\renewcommand{\varDate}{19.-21.11.2025}

\usepackage[utf8]{inputenc} 
\usepackage[T1]{fontenc} 
 
\usepackage{amsmath,amssymb,amsfonts,mathtools,amsthm} 

\usepackage[babel,autostyle]{csquotes}

\usepackage{tabularx}
\newcolumntype{L}[1]{>{\raggedright\arraybackslash}p{#1}} 
\newcolumntype{C}[1]{>{\centering\arraybackslash}p{#1}} 
\newcolumntype{R}[1]{>{\raggedleft\arraybackslash}p{#1}} 
\usepackage{booktabs}

\usepackage{paralist}   

\usepackage[noend]{algpseudocode}

\algrenewcommand\algorithmicrequire{\textbf{Voraussetzung:}}
\algrenewcommand\algorithmicensure{\textbf{Abschlussbedingung:}}

\usepackage{listings} 
\usepackage{subcaption}  

\usepackage{epstopdf}
\usepackage{xcolor}
\selectcolormodel{gray}  

\usepackage{scrhack}
\usepackage{varwidth}
\usepackage{rotating} 
\usepackage[defaultlines=2,all]{nowidow}  
\usepackage{import}
\usepackage{placeins}
\usepackage{makecell}


\usepackage{siunitx} 

\usepackage{amsmath,amssymb,amsfonts,mathtools,amsthm} 

\usepackage{longtable} 
\usepackage{multirow}
\usepackage{tikz}
\usepackage{pgfplots}
\usepackage{smartdiagram}
\usetikzlibrary{patterns}
\usetikzlibrary{shapes} 
\usetikzlibrary{shapes.geometric, arrows,positioning}
\usepackage{smartdiagram}
\usesmartdiagramlibrary{additions} 
\AtBeginDocument{
}
\usepackage{xcolor}
\selectcolormodel{gray}

\usepackage[acronym]{glossaries} 
\usepackage{hyphenat}
\usepackage{etoolbox}
\usepackage{printlen}
\usepackage{graphicx}
\usetikzlibrary{calc, positioning, matrix}
\usepackage{standalone}

\newcommand{\newacr}[4][]{\newacronym[
	sort={\ifthenelse{\isempty{#1}}{#2}{#1}},
	]{#2}{#3}{#4}}
\glsdisablehyper

\robustify{\bfseries}

\newacronym{dl}{DL}{Deep~Learning}
\newacronym{dpf}{dpf}{days~post~fertilization}
\newacronym{echa}{ECHA}{European~Chemicals~Agency}
\newacronym[
  shortplural={EUs},
  longplural={European~Unions},
]{eu}{EU}{European~Union}
\newacronym{fet}{FET}{Fish~Embryo~Acute~Toxicity}
\newacronym{hts}{HTS}{High-Throughput~Screening}
\newacronym{hpf}{hpf}{hours~post~fertilization}
\newacronym[
  shortplural={LLMs},
  longplural={Large~Language~Models},
]{llm}{LLM}{Large~Language~Model}
\newacronym{ml}{ML}{Machine~Learning}

\newacronym[
  shortplural={NAMs},
  longplural={New~Approach~Methodologies},
]{nam}{NAM}{New~Approach~Methodology}
\newacronym{rag}{RAG}{Retrieval-
Augmented~Generation}
\newacronym{reach}{REACH}{Registration, Evaluation, Authorisation~and~Restriction~of~Chemicals}
\newacronym{ssl}{SSL}{Self-Supervised~Learning}
\newacronym{xai}{XAI}{Explainable~Artificial~Intelligence}

\begin{document}


\hyphenpenalty=2000

\pagenumbering{roman}
\cleardoublepage
\setcounter{page}{1}
\pagestyle{scrheadings}
\pagenumbering{arabic}

\setnowidow[2]
\setnoclub[2]

\renewcommand{\Title}{Self-Supervised Learning Strategies for a Platform to Test the Toxicity of New Chemicals and Materials}

\renewcommand{\Authors}{Thomas Lautenschlager\textsuperscript{1}, Nils Friederich\textsuperscript{1,2}, Angelo Jovin Yamachui Sitcheu\textsuperscript{1}, Katja Nau\textsuperscript{1}, Ga\"{e}lle Hayot\textsuperscript{2}, Thomas Dickmeis\textsuperscript{2}, Ralf Mikut\textsuperscript{1}}
\renewcommand{\Affiliations}{\textsuperscript{1}Institute for Automation and Applied Informatics (IAI)\\
	\textsuperscript{2}Institute for Biological and Chemical Systems - Biological Information Processing (IBCS-BIP)\\
    Karlsruhe Institute of Technology\\
	E-Mail: thomas.lautenschlager@kit.edu}

							 
\renewcommand{\AuthorsTOC}{Thomas Lautenschlager, Nils Friederich, Angelo Jovin Yamachui Sitcheu, Katja Nau, Ga\"{e}lle Hayot, Thomas Dickmeis, Ralf Mikut} 
\renewcommand{\AffiliationsTOC}{Karlsruhe Institute of Technology} 

\setLanguageEnglish
							 
\setupPaper 

\section*{Abstract}
High-throughput toxicity testing offers a fast and cost-effective way to test large amounts of compounds. A key component for such systems is the automated evaluation via machine learning models. In this paper, we address critical challenges in this domain and demonstrate how representations learned via self-supervised learning can effectively identify toxicant-induced changes. We provide a proof-of-concept that utilizes the publicly available EmbryoNet dataset, which contains ten zebrafish embryo phenotypes elicited by various chemical compounds targeting different processes in early embryonic development. Our analysis shows that the learned representations using self-supervised learning are suitable for effectively distinguishing between the modes-of-action of different compounds. Finally, we discuss the integration of machine learning models in a physical toxicity testing device in the context of the TOXBOX project.


\section{Introduction}
\begin{figure}[tb]
	\centering
	\includegraphics[width=0.95\textwidth]{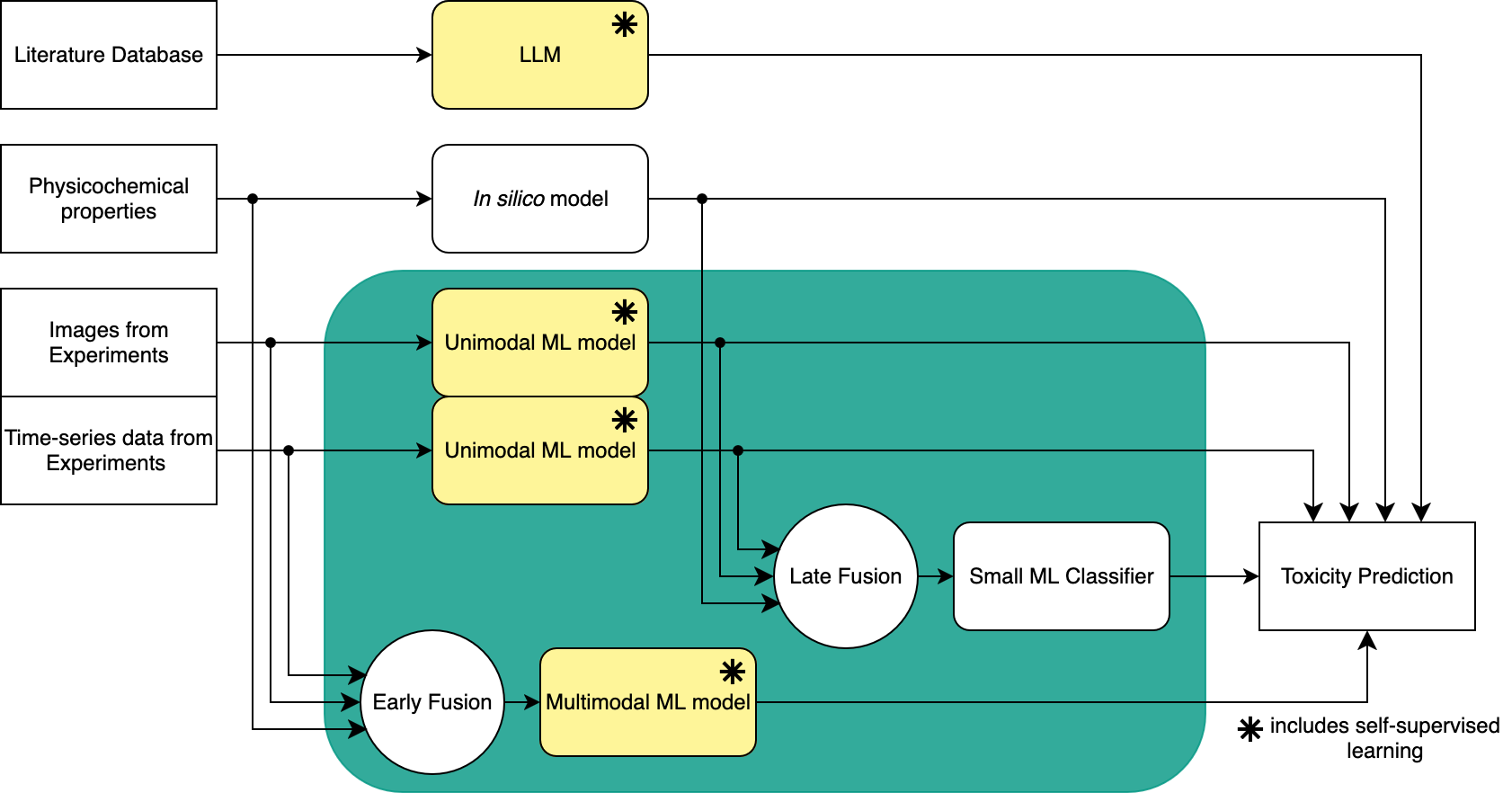}
	\caption{Diagram of different computational models for toxicity predictions. The green box denotes the approaches more closely discussed in this paper. Models that incorporate \gls{ssl} are marked with an asterisk and highlighted in yellow.}	
    \label{fig:toxicity_models}
\end{figure}

The \acrshort{reach} (\acrlong{reach}) regulation, introduced in 2007, aims to better understand chemical compounds entering the \acrshort{eu} (\acrlong{eu}) market \cite{reach}. According to \acrshort{reach}, companies that import or produce a certain compound in quantities exceeding one tonne, are obligated to test the compounds for toxicity and report the results to the \gls{echa} \cite{reach}. Over \SI{23000}{} different compounds were registered under \acrshort{reach} as of 2025 \cite{reachstatistics}.\\
This illustrates the need for large amounts of toxicity tests to be conducted. Typically, these tests are done \textit{in vivo} using rats or other mammals \cite{alternatives}. However, they are relatively costly due to housing and feeding needed, as well as the comparably low reproduction rate of these animals. Furthermore, animal testing requires lengthy legal procedures and poses ethical concerns. Russell and Burch formulated the 3R principles that aim to replace, reduce and refine tests conducted on animals, where possible \cite{3R}. Consequently, interest in alternative forms of toxicity testing is increasing \cite{alternatives}.\\
For approaches deviating from traditional \textit{in vivo} studies, the umbrella term of \glspl{nam} was coined. In the context of \glspl{nam}, tests that are suitable for \gls{hts} are often discussed. These include for example \textit{in vivo} studies using zebrafish (\textit{Danio rerio}) embryos \cite{zebrafish}. 
They are cheaper to rear due to lower maintenance costs and a higher progeny number than animals traditionally used in toxicity testing. Furthermore, according to \acrshort{eu} legislation, zebrafish embryos are not considered animals up until 5 \gls{dpf}, facilitating easier adoption for testing \cite{legal}. \textit{Daphnia magna} is another species that is often considered for \glspl{nam} and \gls{hts} \cite{daphnia}. Apart from \textit{in vivo} testing, \textit{in vitro} tests using cell-based assays or organ models are also rising in popularity \cite{alternatives}. While cell-based assays are suitable for \gls{hts} experiments, the viability of organ models for \gls{hts} is being actively investigated \cite{organoidhts}. More complex approaches, such as organs-on-a-chip or extensions using multiple connected organs for body-on-a-chip systems, are discussed as well \cite{organonachip}.\\
The vast amount of data generated by such \gls{hts} approaches, however, necessitates the use of automated evaluation methods, which can be achieved using \gls{ml} models. While most of the literature on \gls{ml} in toxicology is focused on \textit{in silico} models, research on \gls{ml} for toxicity test automation is fairly scarce \cite{insilico}. However, since \textit{in silico} \gls{ml} models often show poor generalization to compounds with dissimilar properties to the ones they were trained on \cite{insilico}, there is a need for automatic evaluations of experimental \gls{hts} data, using \gls{ml}. The data generated through \gls{hts} is often high-dimensional, encompassing microscopic images and time-series data such as electrochemical readouts. Since \gls{dl} models generally perform better on high-dimensional data than traditional \gls{ml} models \cite{dlbook}, they are better suited for evaluating \gls{hts} data.\\
Incorporating \gls{ssl} into \gls{dl} models can offer various advantages in the domain of toxicity testing. Since labeled data is often scarce, self-supervised pretraining is a valuable technique for building robust models using smaller datasets for downstream tasks such as classification. Furthermore, the continuous representations learned by \gls{ssl} can model concentration-dependent gradients of toxicant-induced changes, while the inherent clustering of the learned representations can identify compounds with similar modes-of-action.\\
Recently, the use of \glspl{llm} has also been discussed in the context of toxicity predictions \cite{llm}. \glspl{llm} are mainly considered in the context of data extraction and data curation from different toxicological databases or from scientific literature \cite{llmdataextraction}. They could also be utilized to directly make predictions based on a given literature database using \gls{rag} or fine-tuned \glspl{llm} \cite{llmfinetune}. However, since \glspl{llm} are prone to hallucinations \cite{llmhallucinations}, their application and wider adoption should be done cautiously. Fusion approaches for \gls{ml} models, combining different inputs, can also be explored. In this way, different experimental data as well as physicochemical properties of the tested compounds could be combined for a single toxicity prediction. A summary of the discussed approaches to computational models for toxicity predictions is given in Figure \ref{fig:toxicity_models}.\\
While the discussion of individual toxicological endpoints is beyond the scope of this paper, it is important to note that computational models often only make predictions on one or a few toxicological endpoints. \gls{ml} models can only accurately predict toxicity for the endpoints for which the model was trained on.\\
In this paper, we mostly focus on discussing image-based approaches because most of the published work on toxicity tests that are viable for \gls{hts} focuses on the automatic evaluation of images. The outline for the rest of this paper is as follows: In Section \ref{relatedwork}, we discuss existing \gls{dl} approaches that are suitable for \gls{hts} scenarios. We provide a brief introduction to \gls{ssl} in general and portray advantages of \gls{ssl} specific to toxicity testing in Section \ref{self-supervised}. Section \ref{proofofconcept} contains a proof-of-concept demonstrating that representations learned by \gls{ssl} can successfully identify toxicant-induced changes and relate the same modes-of-action to each other. Using various analyses, we investigate properties of the learned representations in detail and discuss them. Lastly, in Section \ref{toxbox}, we address challenges arising from the integration of \gls{dl} models into TOXBOX, a real-world toxicity testing device and outline strategies to tackle them.


\section{Related Work}
\label{relatedwork}
\begin{sloppypar}
There are several works concerned with using \gls{dl} models for evaluating toxicity tests, which we are going to discuss here. These works could be adapted for use in \gls{hts} scenarios. However, regarding species commonly used in toxicity testing, there is a notable lack of research that consistently focuses on a single aspect of toxicity testing for a specific species. Even less work focuses on the automation of established toxicity tests.\\
As already mentioned, zebrafish are a model organism often discussed in the context of toxicity testing \cite{zebrafish}. Several test protocols have been established for zebrafish embryos and larvae <5 \gls{dpf}. For example, the \gls{fet} Test is OECD-approved \cite{fet} and is often discussed as an alternative to the Fish Acute Toxicity Test \cite{aft}, which uses adult fishes and is therefore not suitable for \gls{hts}. Various behavioral tests have also been established. These tests, though often lacking standardized protocols, enable the evaluation of neurotoxicity \cite{zebrafishbehavioral}.\\ 
\gls{dl} techniques for the classification of abnormally developing zebrafish embryos have been proposed in several studies \cite{morphologicalattention, morphological1, morphological2, embryonet}.
However, the different investigations show low consistency. Most of these focus on the classification of hatched eleutheroembryos during different timepoints \cite{morphologicalattention, morphological1, morphological2}, only one publication focuses on earlier embryonic stages \cite{embryonet}. All the publications use different classes in their classification approaches. These inconsistencies in the existing research make comparisons difficult.\\ 
There are also no publications that tackle the automatic evaluation of the \gls{fet}, the only OECD-approved toxicity test involving zebrafish <5 \gls{dpf} that focuses on phenotypical changes \cite{fet}. However, the automatic identification of coagulated zebrafish embryos, one of the endpoints of the \gls{fet}, has been addressed in several works \cite{embryonet, alshut}. Existing approaches have also paid little attention to identifying modes-of-action of the compounds. Only \v{C}apek et al. \cite{embryonet} define the classified phenotypes based on different developmental pathways that can be blocked by certain toxicants. These phenotypes, however, also do not cover all possible toxicant-induced changes.\\
Several papers focus on \gls{dl} approaches for toxicity testing using \textit{Daphnia magna}. They include models that determine and quantify morphological changes in \textit{Daphnia magna} due to toxicant exposure \cite{daphniamorphological}, models to determine the size and growth rate \cite{growth} and approaches for tracking \textit{Daphnia magna} \cite{tracking} as well as identifying compounds based on locomotor tracks \cite{trackingclassification}.\\
Few approaches exist for the use of \gls{dl} models in cell assays or organ models for toxicity testing. One approach automatically detects the nuclei of the cells and classifies them as either 'healthy' or 'toxicity-affected' \cite{toxrcnn}. Another approach uses time-series data based on a cell impedance signal for the classification of different modes-of-action \cite{impedance}. Cell tracking approaches such as \cite{scherr, embedtrack} are also suitable for toxicity testing, since features such as the number of cells or size of cells can also be used to make predictions on the toxicity of a certain compound.\\ 
Hu et al. \cite{skinmodel} use \gls{dl} models for predicting the thickness of a skin model. They show that lower thickness of the epidermal layer can be used to predict skin toxicity.\\
To the best of our knowledge, only two studies are using \gls{ssl} that can be considered for the automatic evaluation of toxicity tests. Toulany et al. \cite{uncovering} use a Twin Network trained with a triplet loss to investigate the embryonic development in zebrafish. The trained network can be used to determine the similarity between embryo images. This is used for identifying different developmental stages, comparisons regarding the development of zebrafish embryos under different temperatures and detecting deviations from normal development. The authors show that the model can also identify deviations from normal development that are toxicant-induced \cite{uncovering}.\\
In the second paper on \gls{ssl} in toxicity testing, Gendelev et al. \cite{selfsupervisedzebrafish} use Twin Networks on the Motion Index, a measurement of movement based on pixel intensity changes between frames in videos, from different behavioral tests using 7 dpf zebrafish larvae. This approach can group similar modes-of-action together.
\end{sloppypar}


\section{Potential of Self-Supervised Learning in Toxicology} 
\label{self-supervised}
\subsection{Overview of Self-Supervised Learning}
\begin{figure}[tb]
	\centering
	\includegraphics[width=0.95\textwidth]{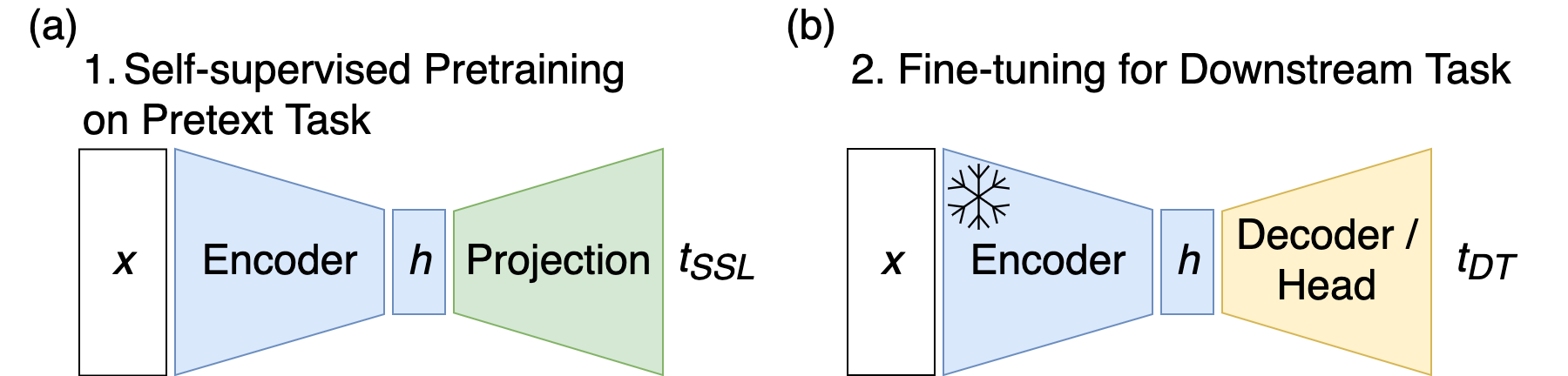}
	\caption{Simplified diagram of (a) pretraining via \gls{ssl} and (b) fine-tuning for downstream task. The encoder is usually frozen during fine-tuning. $x$ denotes the input of the model, $h$ the learned latent space, and $t_{SSL}$ and $t_{DT}$ the targets of the self-supervised learning task and the downstream task, respectively.}
    \label{fig:ssl}
\end{figure}

\gls{ssl} encompasses methods that use a pretext task for learning useful lower-dimensional representations \cite{surveyselfsupervised2}. The pretext task focuses on optimizing for a target $t_{SSL}$ that can be generated from the data itself \cite{surveyselfsupervised}. Depending on the self-supervised algorithm, the pretext tasks in computer vision can range from predicting the correct order of shuffled image patches \cite{jigsaw}, mapping two differently augmented views of the same image together \cite{simclr}, or reconstructing image patches that were masked in the input \cite{mae}.\\ 
Typically, \gls{ssl} uses an encoder that maps the inputs $x$ to a latent space $h$. Often, some kind of projection is used on the latent space $h$, the output of which is used for the optimization regarding the target $t_{SSL}$. This can, for example, be a linear layer \cite{moco}, a projection head \cite{simclr} or a decoder \cite{mae}.\\
The desired output of the \gls{ssl} models are the lower-dimensional representations of the data in the learned latent space $h$. Generally, in the latent space $h$, representations from similar inputs are mapped closely together, while representations from dissimilar inputs are mapped further away from each other. In computer vision, these representations are often referred to as visual representations. For the sake of brevity and because \gls{ssl} can also be used to attain lower-dimensional representations from non-image data, we use the term 'representations' for the rest of this paper.\\
In general computer vision tasks, \gls{ssl} is often used for pretraining \cite{surveyselfsupervised2}. The learned representations are then used to fine-tune for a specific downstream task, such as classification or segmentation \cite{surveyselfsupervised2}. Depending on the kind of task, a decoder or head is used on top of the usually frozen encoder. The decoder or head is then trained regarding the target $t_{DT}$ of the downstream task. This typical training procedure is pictured in Figure \ref{fig:ssl}. Popular \gls{ssl} methods for images include: SimCLR \cite{simclr}, MoCo and its extensions \cite{moco, mocov2, mocov3}, BYOL \cite{byol}, SwAV \cite{swav}, DINO and its extensions \cite{dino, dinov2, dinov3}, Masked Autoencoders \cite{mae} and SimSiam \cite{simsiam}.
\subsection{Label-Efficient Model Training}
Using pretrained models for training on downstream tasks can also be applied to toxicity testing. This allows for the training of more label-efficient models. Further, the representations learned by \gls{ssl} are often more generalizable, enabling transfer learning, where the representations from a different, but similar dataset can be used for fine-tuning to a task, where both data and labels are scarce.\\
Since few image datasets for toxicity testing are published and generating new datasets is expensive, leveraging similar datasets in \gls{ssl} pretraining makes research on \gls{ml} models for toxicity testing more feasible. A strategy that could be used to identify suitable datasets was outlined by Yamachui Sitcheu et al. \cite{similardatasets}. 
\subsection{Continuous Representations}
The latent space $h$ learned by \gls{ssl} methods offers several additional advantages for toxicity testing. A problem when applying supervised \gls{dl} models to toxicity tests is that the methods often used do not account for properties specific to toxicity testing.\\
For example, classification often falls short when evaluating toxicant-induced morphological changes. These changes are usually continuous and the cutoff point is often based on observer experience, with little to no standardization. Thresholds of toxicant-induced changes can therefore vary between studies. It is often unclear whether small changes are already labeled as 'toxic' or only if a clear abnormal phenotype can be observed. Additionally, there are studies that use several classes \cite{embryonet, daphniamorphological} for different magnitudes of the same phenotypic changes, resulting in even more hazily defined cutoff points. By mapping the samples into a latent space that allows for continuous representations of the morphological changes, \gls{ssl} offers an elegant solution for this problem. The resulting representation not only allows fine differentiation based on phenotypic changes but can be thought of as concentration-dependent gradients. The learned representations of a certain phenotypical change will be mapped into the same direction away from healthy phenotypes. Since toxicant-induced changes get more severe with higher concentrations of the compound, the distance to the healthy phenotypes will increase with higher concentrations of the compound.\\
Another downside of supervised classification is that biases can be introduced if small changes due to a toxicant are not represented in the labels used for training. For example, the EmbryoNet-Prime \gls{dl} model, trained on data with labels shifted 4 hours into the past, can identify morphological changes earlier than the expert who labeled the data \cite{embryonet}, indicating that small phenotypical changes are already present.\\
However, simply shifting the labels can result in false labels, since it is unclear when the phenotypical changes first occur. This can deteriorate model performance. \gls{ssl} could potentially identify when small phenotypical changes occur without label-induced biases.
\subsection{Identification of Similar Modes-of-Action}
The clustering inherent to \gls{ssl} can also be used for the identification of similar modes-of-action. Since similar images will be mapped together and dissimilar images mapped away from each other, similar phenotypes induced by compounds with a similar mode-of-action will be clustered. Gendelev et al. \cite{selfsupervisedzebrafish} have shown that this is possible using time-series data of pixel intensity changes from behavioral zebrafish tests.\\
Additionally, a classifier without rejection class forces unknown toxicant-induced changes into one of the classes known from the training dataset. In the case of \gls{ssl}, the representations of an unknown morphological change are mapped away from the representations of the known classes, making it apparent that the representations do not belong to any of the known classes.


\section{Preliminary Experiment: Proof-of-Concept}
\label{proofofconcept}
\subsection{Methods}
We chose SimCLR \cite{simclr} for our proof-of-concept investigating the properties of the latent space of a \gls{ssl} model trained on image data showing toxicant-induced morphological changes. SimCLR is an important baseline in the field of \gls{ssl} and learns meaningful representations of the data. The self-supervised target $t_{SSL}$ of SimCLR aims at minimizing the distance between the representations of two differently augmented views of the same image while maximizing the distance between views of different images. SimCLR uses cosine similarity as the distance measure between different views \cite{simclr}. The resulting latent space is a hypersphere on which the representations are mapped.\\
We used ResNet50 \cite{resnet} as the backbone, which we trained using SimCLR \cite{simclr}. After the training, we evaluated the latent space using linear probing, where a linear classifier is trained on the representations that the frozen backbone outputs. This method is a standard procedure in \gls{ssl} research to assess the quality of learned representations.\\
Since SimCLR training is unsupervised, the labels of the dataset are only used for training the linear classifier. This usually results in worse performance than training the network fully supervised. However, it can still be useful to evaluate the success of self-supervised training.\\
To better understand the latent space and the representations SimCLR learned, we visualize the latent space using UMAP for dimensionality reduction \cite{umap}.\\ 
Furthermore, we investigate the representations of each class. Through a forward pass using the training dataset, we obtain the representations for the training dataset. We calculate the centers of each class by taking the mean of the respective class representations
\begin{equation}
	\pmb{c}_k=\frac{1}{\vert C_k\vert}\sum_{i\in C_k}\pmb{h}_i
	\label{eq:centercalculation}
\end{equation}
where $\pmb{h}_i$ denotes one learned representation, $\pmb{c}_k$ represents the center $\pmb{c}$ for class $k$, $C_k$ denotes the set of representations belonging to class $k$ and $\vert C_k \vert$ its cardinality. The dimensionality of $\pmb{h}$ and $\pmb{c}$ are dependent on the type of network used. Since we use ResNet50 as a backbone, the resulting dimensionality for $\pmb{h}$ and $\pmb{c}$ is 2048 in our analyses.\\
After the calculation, the centers are normalized
\begin{equation}
	\tilde{\pmb{c}}_k=\frac{\pmb{c}_k}{\| \pmb{c}_k\|}
	\label{eq:normalization}
\end{equation}
where $\tilde{\pmb{c}}_k$ refers to the normalized class center of class $k$ and $\| \pmb{c}_k\|$ to the Euclidean norm of class center $\pmb{c}_k$.\\
Next, the class centers are used to calculate the mean cosine similarity of the representations of each class to each center
\begin{equation}
	\overline{\mbox{sim}_{\cos}}(C_l;\space \tilde{\pmb{c}}_k) = \frac{1}{\vert C_l \vert}\sum_{i \in C_l}\pmb{h}_i^\top \cdot \tilde{\pmb{c}}_k
	\label{eq:distancecenterrepresentation}
\end{equation}
where $\overline{\mbox{sim}_{\cos}}(C_l;\space \tilde{\pmb{c}}_k)$ denotes the mean cosine similarity between the set $C_l$ that includes the representations $\pmb{h}_i$ of the test dataset. Note that the representations of $\pmb{h}_i$ and $\tilde{\pmb{c}}_k$ are both normalized in this calculation.\\
The distance of a point to a class center can be thought of as an anomaly score for that class. To achieve a deeper understanding of the constructed latent space, we also calculate the cosine similarities between the different class centers:
\begin{equation}
	\mbox{sim}_{cos}(\tilde{\pmb{c}}_l; \space\tilde{\pmb{c}}_k) = \tilde{\pmb{c}}_l^\top \cdot \tilde{\pmb{c}}_k
	\label{eq:distancecentercenter}
\end{equation}


\subsection{Dataset}
For our analyses, we used the publicly available EmbryoNet dataset \cite{embryonet}. It features images of ten different zebrafish embryo phenotypes. There are seven phenotypes, where the used toxicant targeted a major signaling pathway in early embryonic development. The respective phenotypes are named after their affected pathway and whether a loss-of-function or gain-of-function is present: -BMP, +RA, -Wnt, -FGF, \mbox{-Nodal}, -Shh and -PCP. Other classes include the 'Normal' class, featuring normally developing embryos, the 'Dead' class, which includes embryos that have died and coagulated, and the 'Unknown' class, for embryos whose phenotype could not be identified.\\
The original dataset features embryos that are periodically imaged from 2 \gls{hpf} to 26 \gls{hpf}, since developmental aspects are closely discussed in the EmbryoNet paper \cite{embryonet}. However, since we were most interested in classifying the different phenotypes that only become apparent as development progresses, we chose to use only images from the later timepoints. This not only reduced training time but also avoided learning visual features that are not necessary for phenotype classification.\\ 
The EmbryoNet dataset consists of images of wells containing multiple zebrafish embryos \cite{embryonet}. We adopted the predefined split between training and test dataset of the Embryo\-Net dataset. Additionally, we defined a validation dataset using images from 10\% of the wells that make up the training dataset. The chosen wells were randomly sampled. For extracting the individual embryo crops from the well images, we used the bounding boxes, which are provided with the dataset. For the training and validation dataset we used the crops of embryos ranging from 25 \gls{hpf} to 26 \gls{hpf}. The evaluations were done using only the last crop of each embryo in the test dataset, which was recorded at 26 \gls{hpf}. The resulting training, validation and test datasets consist of \SI{135475}{}, \SI{15670}{} and \SI{772}{} embryo crops, respectively.


\subsection{Implementation Details}
\label{implementation}
\begin{sloppypar}
We used MMPretrain \cite{mmpretrain} for training SimCLR and the linear classifier. The training was done using 8 NVIDIA A100-40s. Other evaluations were done using custom code and were run on an NVIDIA RTX 3090. The code is available at \url{github.com/lautthom/self_supervised_learning_strategies_toxicity_testing}. Details regarding the hyperparameters and augmentations used are available in the config files used for MMPretrain provided with the code.\\
Unless otherwise noted, we used the same hyperparameters as described in the original SimCLR paper \cite{simclr}. We trained SimCLR for 200 epochs and reduced the batch size to 2048. The learning rate was adjusted accordingly with square root scaling \cite{squarerootscaling}. \\
For the linear classifier, we also adopted the training and testing procedure as well as the hyperparameters as defined in the SimCLR paper \cite{simclr}, unless otherwise noted. Since the classes in the EmbryoNet dataset are imbalanced, we used a weighted loss function for the training of the linear classifier
\begin{equation}
	w_i=\frac{n}{n_i \cdot C}
	\label{eq:weightcalculation}
\end{equation}
where $w_i$ specifies the weight of class $i$ in the loss function, $n$ the number of total samples, $n_i$ the number of samples in class $i$ and $C$ the number of classes. Furthermore, we used early stopping based on the accuracy the linear classifier achieved on the validation dataset.\\
The augmentations were adjusted to fit the domain-specific needs of zebrafish embryo images. The following augmentations were used for both the SimCLR training and the linear classifier training: random crop, horizontal flip, rotations of up to 360°, random brightness changes, random contrast changes, CLAHE, sharpen, motion blur, defocus, grid distortion, optical distortion, elastic transform, salt and pepper noise, Gaussian noise, Poisson noise and solarize.
\end{sloppypar}


\subsection{Results}
\label{results}
\begin{figure}[h!]
	\centering
	\includegraphics[width=0.95\textwidth]{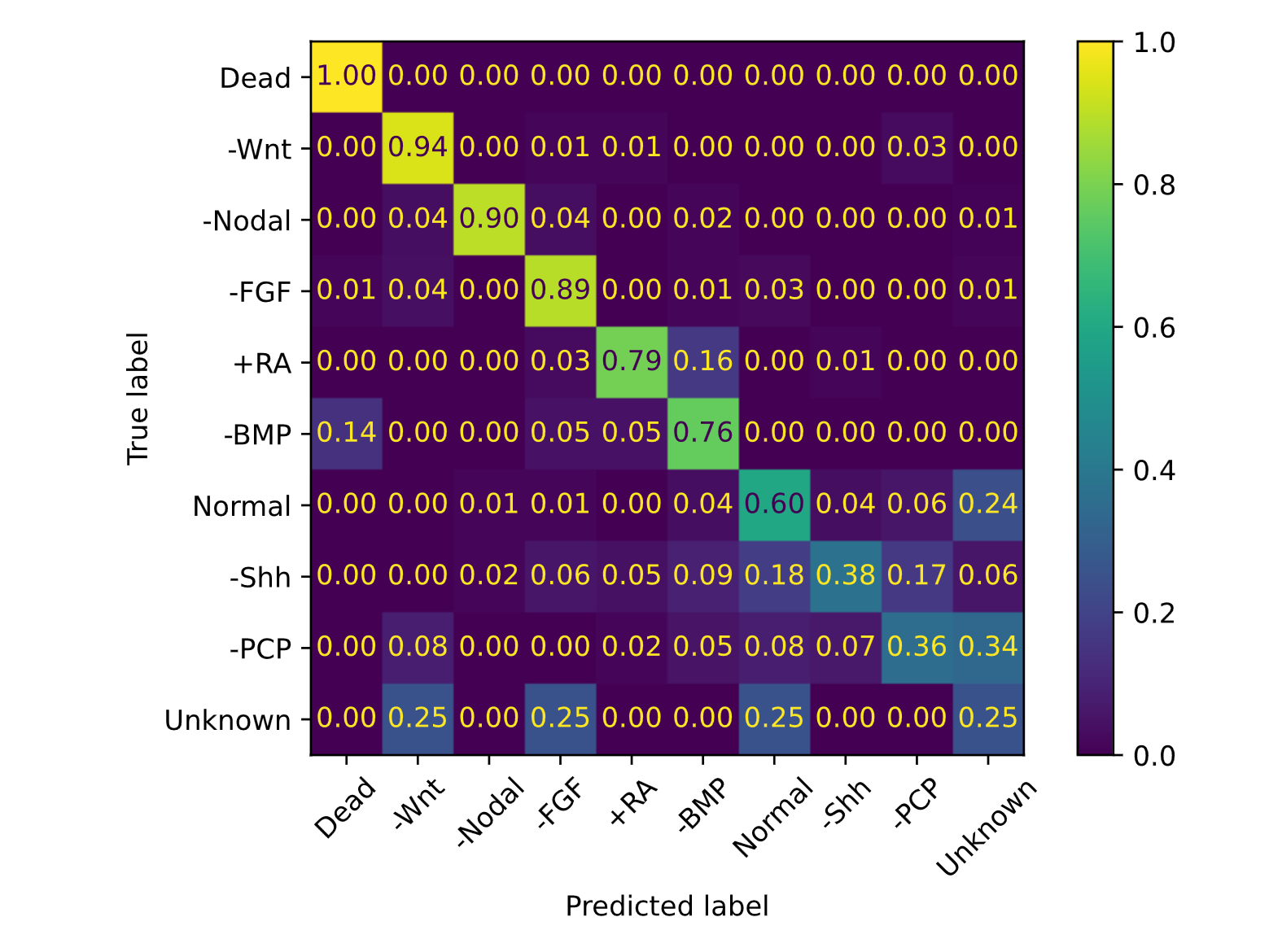}
	\caption{Confusion matrix of the linear classifier trained using SimCLR representations}
	\label{fig:confusionmatrix}
\end{figure}

\begin{figure}[h!]
	\centering
	\includegraphics[width=0.95\textwidth]{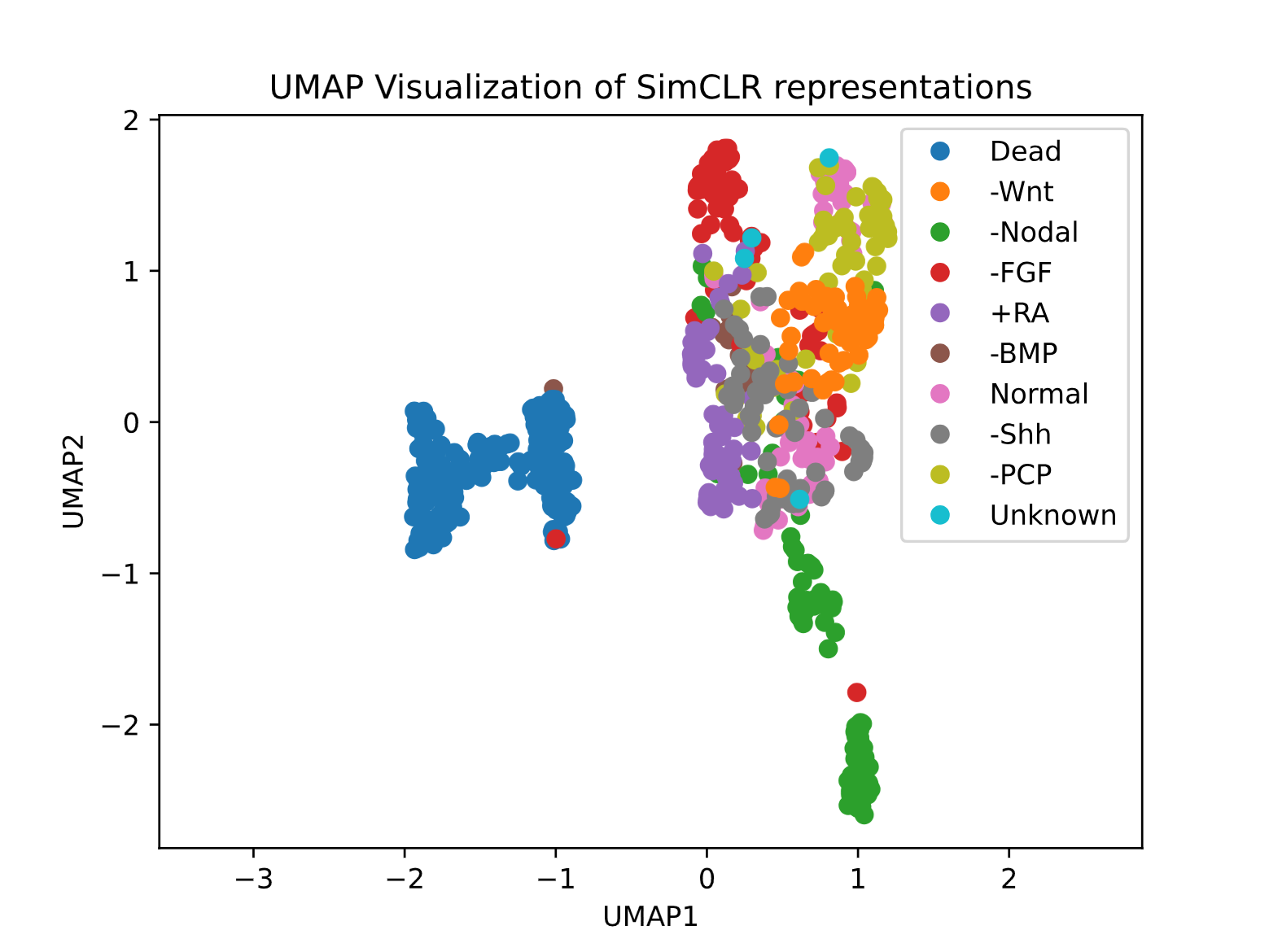}
	\caption{UMAP visualization of SimCLR representations}
	\label{fig:umap}
\end{figure}

\begin{figure}[h!]
	\centering
	\includegraphics[width=0.95\textwidth]{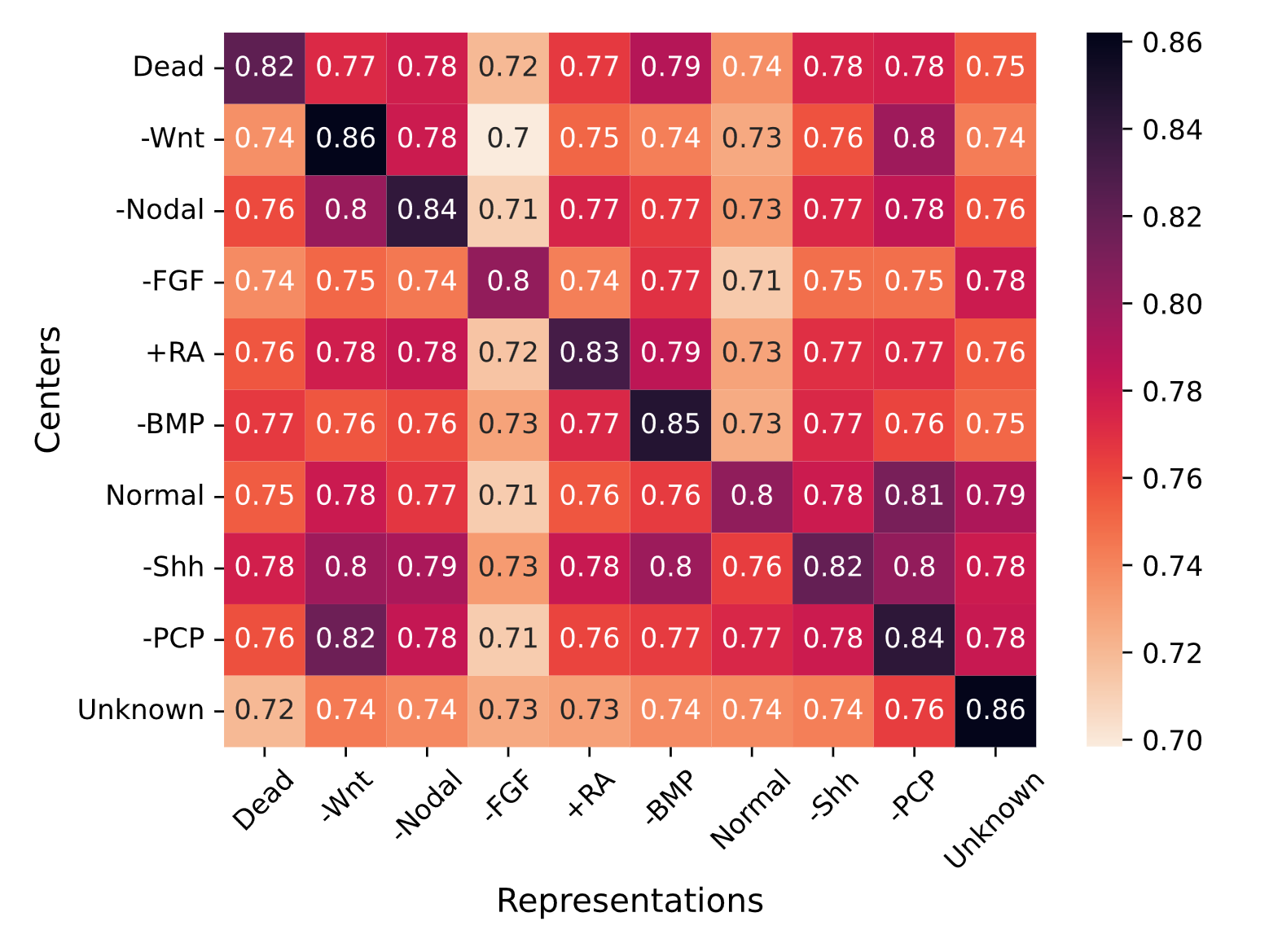}
	\caption{Mean cosine similarities between class centers and representations of the class}
	\label{fig:distancecentersrepresentations}
\end{figure}

\begin{figure}[h!]
	\centering
	\includegraphics[width=0.95\textwidth]{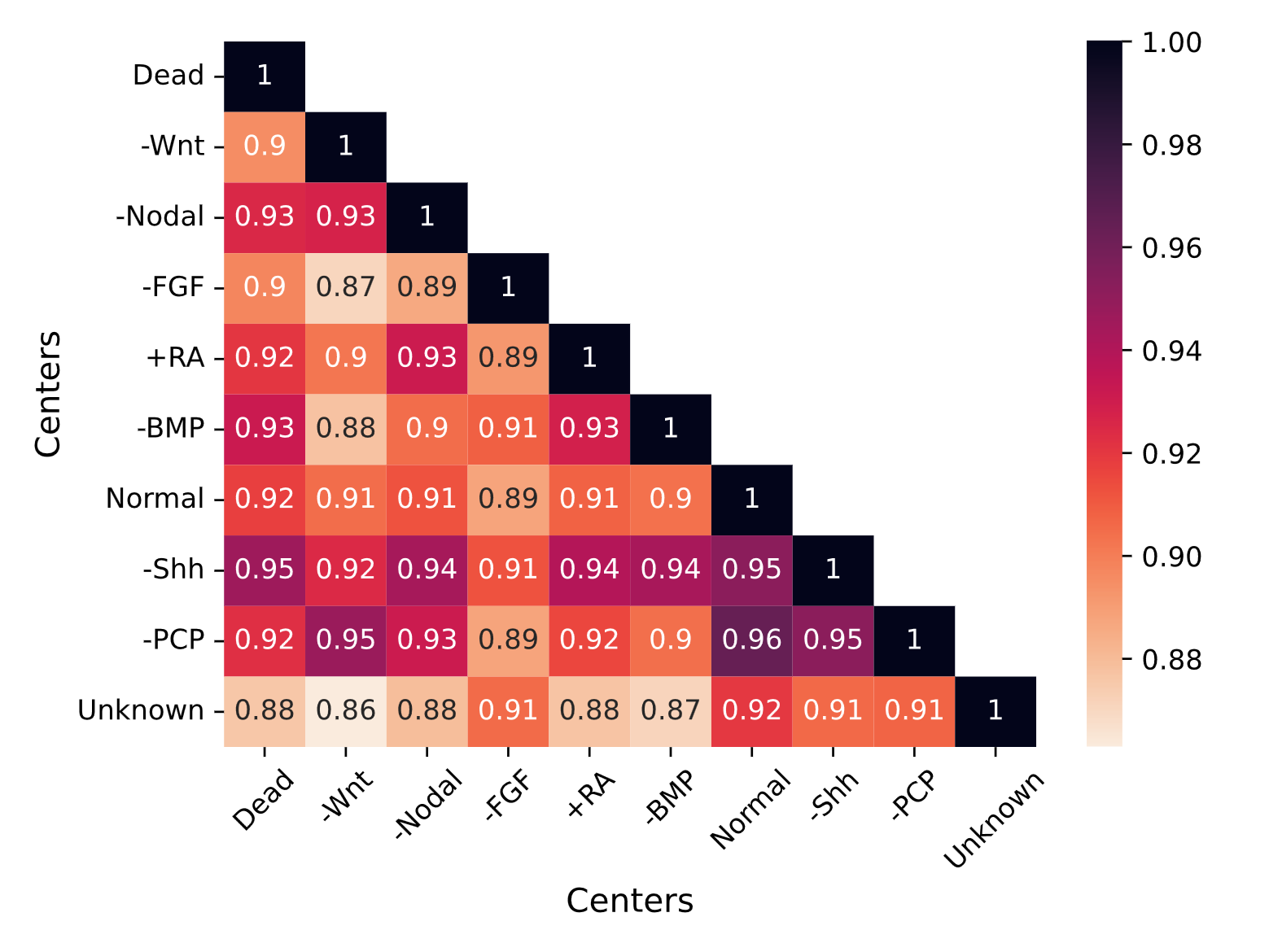}
	\caption{Cosine similarities between the class centers}
	\label{fig:distancecenters}
\end{figure}

The linear classifier trained on top of the representations learned by SimCLR achieved an accuracy of 79.9\% on the test dataset. This is about ten percentage points below the 89\% accuracy reported in the original EmbryoNet paper \cite{embryonet}. The normalized confusion matrix of the linear classifier trained on the 10 classes is depicted in Figure \ref{fig:confusionmatrix}. Dead embryos are classified most reliably with a recall of 100\%. Other classes with a high recall are the -BMP, -FGF, -Nodal, +RA and -Wnt phenotypes. The 'Normal' class has a relatively low recall of 60\%. The -PCP and -Shh phenotypes also have a low recall. The 'Unknown' phenotype has the lowest recall, however, only 4 images belong to the 'Unknown' class in our test dataset.\\
An UMAP visualization of the learned representations of the test dataset is given in Figure \ref{fig:umap}. The visualization reflects the recall values given in Figure \ref{fig:confusionmatrix}. The 'Dead' class is mapped far away from the other classes. The representations of the other classes with a high recall are also mapped close to each other and are fairly easy to distinguish from the representations of other classes.\\
For each class, we calculated the mean cosine similarity between the representations and the centers of the respective class. The results are given in Figure \ref{fig:distancecentersrepresentations}. The lowest mean cosine similarity is 0.70. However, for all classes, the mean cosine similarities between their representations and their respective class centers are higher than the mean cosine similarities to all other class centers.\\
Figure \ref{fig:distancecenters} shows the cosine similarities between the different class centers. As in Figure \ref{fig:distancecentersrepresentations} all cosine similarities are fairly high. The lowest cosine similarity between two class centers is 0.86.\\
Due to the high cosine similarities between the centers and representations in Figure \ref{fig:distancecentersrepresentations} as well as between the centers themselves in Figure \ref{fig:distancecenters}, we took a closer look at the cosine similarities of the individual representations. The minimal cosine similarity between two representations is 0.37, while the mean similarity is 0.64 and the highest cosine similarity is 1.0. This means that only a small part of the hypersphere that the images are mapped to during \gls{ssl} training is populated by the representations.


\subsection{Discussion, Limitations and Outlook}
Our investigation of the learned latent space in Section \ref{results} leads to different insights. The performance of the linear classifier shows that the learned representations have a fairly good quality. While the classification is markedly worse than the supervised baseline, it still reaches acceptable performance. Since most of the classes correspond to a certain mode-of-action, the clustering based on modes-of-action works fairly well. However, the phenotypes in the EmbryoNet dataset are elicited by the same compounds, which makes it hard to evaluate if the clustering was indeed based on the mode-of-action or on some other compound-specific properties. Certain phenotypes reach a particular low recall, for example the -Shh and -PCP phenotype. However, this is also true for the supervised model, as well as for the evaluations of experienced developmental biologists, reported in the EmbryoNet paper \cite{embryonet}.\\ 
An interesting approach for future research could be to evaluate the representations learned via \gls{ssl} on other downstream tasks, such as segmentation or transfer learning to another classification task. Presumably, the representations learned by \gls{ssl} should outperform the ones learned during supervised training in these tasks. A possible application would be the automatic evaluation of the \gls{fet} \cite{fet}. Future investigations could also explore fine-tuning the models with fewer available labels and compare the performance deterioration to fully supervised learning.\\
A problem with the present latent space is that the representations only populate a small area of the hypersphere. Investigations show that more uniform distributions on the hypersphere generally improve performance \cite{hypersphereuniform}. It is unclear whether this is also true for domain-specific use cases, where the images are very similar to each other.\\
Further investigations are needed to see if the theoretical considerations presented in Section \ref{self-supervised} can be empirically verified. Clustering based on similar modes-of-action was already shown to be feasible for one other application \cite{selfsupervisedzebrafish} and the results presented in this paper support this finding. Given this, it is likely that concentration-dependent gradients in the latent space also exist. Unfortunately, the EmbryoNet dataset is not best suited for this investigation, since the authors used the same concentrations for the elicitation of most of the phenotypes \cite{embryonet}.


\section{Integration with TOXBOX device}
\label{toxbox}
\begin{figure}[tb]
	\centering
	\includegraphics[width=0.95\textwidth]{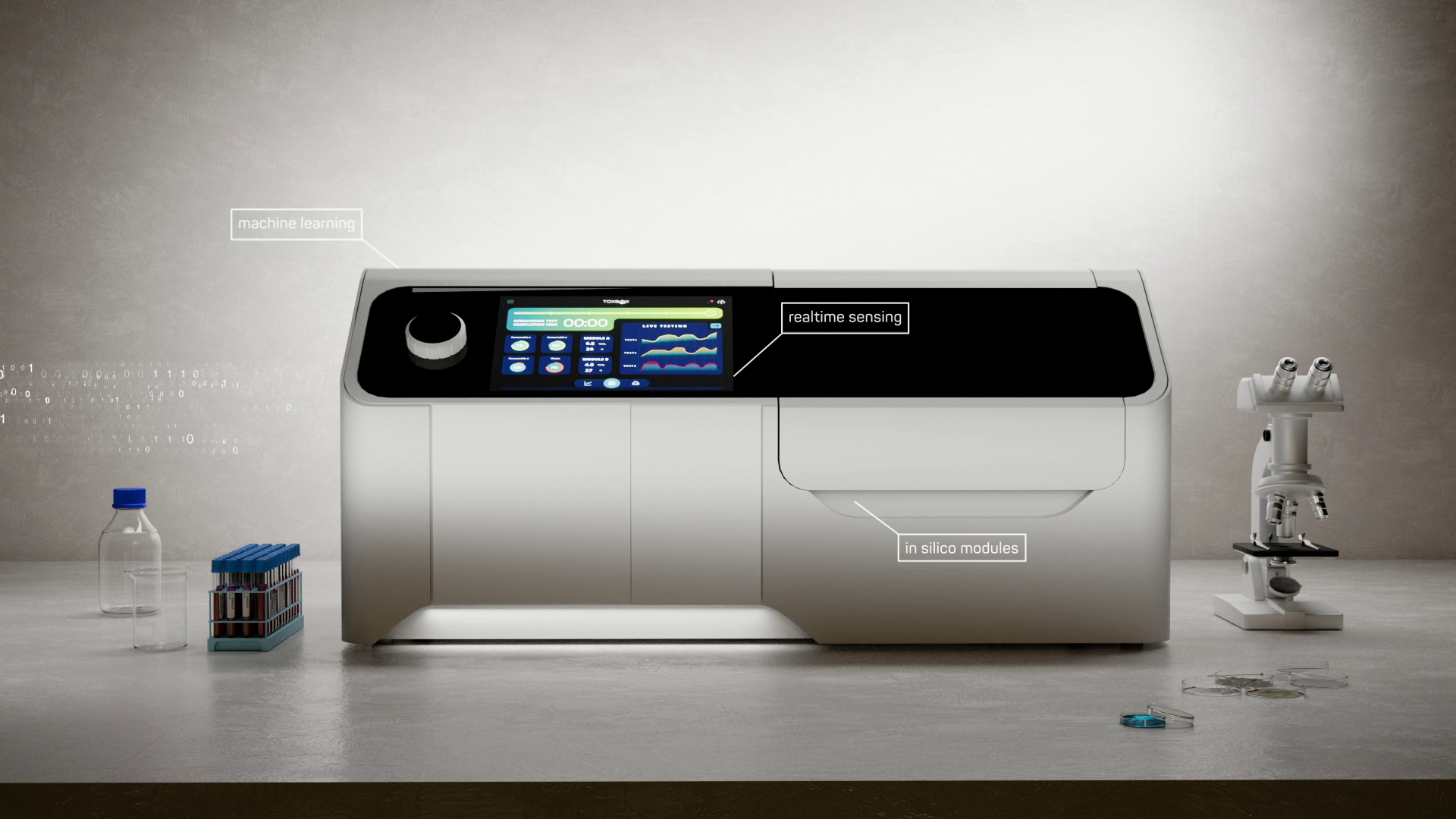}
	\caption{Image of the TOXBOX prototype; screenshot from \cite{toxboxvideo}}
	\label{fig:toxbox}
\end{figure}

Since the aim is to integrate \gls{ml} models in high-throughput processes, the \gls{ml} models also need to be integrated with a physical toxicity testing device. We discuss this in the context of the TOXBOX\footnote{https://toxbox.eu} project. The project aims to design an all-in-one platform for reliable toxicity testing \cite{toxbox}. A prototype of the TOXBOX device is pictured in Figure \ref{fig:toxbox}.\\
TOXBOX will feature different \textit{in vitro} organ models as well as a zebrafish embryo module \cite{toxbox}. Due to the advantages illustrated in this paper, pretraining the models via \gls{ssl} and then fine-tuning them to the specific toxicity prediction task seems to be the most viable option. This should be especially advantageous if the data and/or labels generated during the TOXBOX project are scarce and data from similar datasets can be leveraged using \gls{ssl}. Fully supervised models should be trained as well and compared to \gls{ssl} models, to ensure that the model with the best performance will be used.\\
\gls{xai} methods can help in the evaluation of the different \gls{ml} models \cite{explainableai}. Since \gls{xai} makes the underlying factors that lead to a certain prediction of a \gls{ml} model more transparent, experts can assess whether the factors used are indicative of toxicity.\\
Furthermore, the latent space of the \gls{ssl} models could be used to gain more information about the tested compound. Based on the distance between the representations of known compounds and the tested compound, it can be determined whether the compound has a similar mode-of-action as known compounds or whether it has an unknown effect. Compounds with unknown effects merit more thorough investigation, both on the compounds themselves and also if something went wrong during testing.\\
A crucial topic to discuss when using \gls{ml} models in toxicity testing is concept drift. Concept drift refers to gradual changes in the underlying data, which happen over time and can deteriorate the model's performance \cite{conceptdrift}. This can happen especially easily in toxicity testing if groups of compounds are tested that differ in important aspects from those used in acquiring the training data, particularly when compounds with different modes-of-action are tested. This means that the model's performance needs to be closely monitored to ensure reliable predictions.\\
If new modes-of-action are found or the model's performance drops due to concept drift, it may be necessary to retrain the model. This can be challenging, as the newly acquired data from the device may be highly imbalanced. Further, the data of the previously unknown mode-of-action can be scarce. Different strategies for retraining the models should be explored and closely evaluated in such scenarios.


\section{Conclusion}
In this paper, we have illustrated how aspects inherent to \gls{ssl} are suitable to address different challenges specific to toxicity testing, specifically dealing with sparse labeled data, accounting for continuous changes due to toxicant exposure and identifying similar modes-of-action. We provided a proof-of-concept that demonstrates how representations learned via \gls{ssl} can, in practice, be utilized for toxicity testing. Further, we discussed various challenges involved in adapting machine learning models to physical toxicity testing devices.


\section{Acknowledgments}
This work has received funding from the European Union's HORIZON-CL4-2023-RESILIENCE-01 Research and Innovation programme under Grant Agreement No 101138387 (TOXBOX). Views and opinions expressed are however those of the authors only and do not necessarily reflect those of the European Union. The European Union cannot be held responsible for them.\\
This work is supported by the Helmholtz Association Initiative and Networking Fund on the HAICORE@KIT partition.\\
The present contribution is supported by the Helmholtz Association under the joint research school "HIDSS4Health - Helmholtz Information and Data Science School for Health." and the program "Natural, Artificial and Cognitive Information Processing (NACIP)".

\subsection{Author contributions}
The authors used the AI language models GPT-4.5 and GPT-5.0 to improve the language and style of this manuscript. The authors have accepted responsibility for the entire content of this manuscript and approved its submission. We describe the individual contributions of T. Lautenschlager (TL), N. Friederich (NF), A.J. Yamachui Sitcheu (AJYS), K. Nau (KN), G. Hayot (GH), T. Dickmeis (TD), R. Mikut (RM): Conceptualization: TL, NF, RM; Methodology: TL, NF, RM; Software: TL; Investigation: TL; Writing - Original Draft: TL; Writing - Review \& Editing: TL, NF, AJYS, KN, GH, TD, RM; Supervision: NF, RM; Project administration: RM; Funding Acquisition: TD, RM.



\addtocontents{toc}{\protect\newpage}



\end{document}